\documentclass[sn-vancouver,Numbered,iicol]{sn-jnl}

\usepackage{graphicx}%
\usepackage{multirow}%
\usepackage{amsmath,amssymb,amsfonts}%
\usepackage{amsthm}%
\usepackage{mathrsfs}%
\usepackage[title]{appendix}%
\usepackage{xcolor}%
\usepackage{textcomp}%
\usepackage{manyfoot}%
\usepackage{booktabs}%
\usepackage{algpseudocode}%
\usepackage{listings}%

\usepackage{times}
\usepackage{helvet}
\usepackage{courier}
\usepackage{csquotes} 
\usepackage{color}
\usepackage{paralist}
\usepackage{indentfirst}
\usepackage{subfigure}
\usepackage{float}

\usepackage{pgfplots} 
\usepackage{epstopdf}
\usepackage{epsfig}
\pgfplotsset{compat=newest}
\definecolor{forestgreen}{RGB}{0,139,69}
\usepackage{array}

\definecolor{SeaGreen4}{RGB}{0,205,102} 
\definecolor{SlateBlue}{RGB}{106,90,205} 
\definecolor{DarkRed}{RGB}{178,34,34} 
\usepackage{pifont}
\newcommand{\cmark}{\ding{51}}%
\newcommand{\xmark}{\ding{55}}%
\usepackage{makecell}
\usepackage{caption}
\usepackage{enumerate}
\usepackage{multirow}
\usepackage{multicol}

\jyear{2023} 

\theoremstyle{thmstyleone}%
\theoremstyle{thmstyletwo}%

\theoremstyle{thmstylethree}%

\begin{document}

\title[COESOT]{Revisiting Color-Event based Tracking: A Unified Network, Dataset, and Metric}

\author[1,2]{\fnm{Chuanming} \sur{Tang}}\email{tangchuanming19@mails.ucas.ac.cn}
\author*[3]{\fnm{Xiao} \sur{Wang}}\email{xiaowang@ahu.edu.cn}
\author[3]{\fnm{Ju} \sur{Huang}}\email{huangju991011@163.com}
\author[3]{\fnm{Bo} \sur{Jiang}}\email{zeyiabc@163.com}
\author[4]{\fnm{Lin} \sur{Zhu}}\email{linzhu@pku.edu.cn}
\author*[1,2]{\fnm{Jianlin} \sur{Zhang}}\email{jlin@ioe.ac.cn}
\author[5]{\fnm{Yaowei} \sur{Wang}}\email{wangyw@pcl.ac.cn}
\author[5,6,7]{\fnm{Yonghong} \sur{Tian}}\email{yhtian@pku.edu.cn}

\affil[1]{\orgname{University of Chinese Academy of Sciences}, \orgaddress{\city{Beijing},  \country{China}}}

\affil[2]{\orgname{Institute of Optics and Electronics, Chinese Academy of Sciences}, \orgaddress{\city{Chengdu},  \country{China}}}
 
\affil[3]{\orgdiv{School of Computer Science and Technology}, \orgname{Anhui University}, \orgaddress{  \state{Hefei}, \country{China}}}

\affil[4]{ \orgname{Beijing Institute of Technology}, \orgaddress{  \state{Beijing}, \country{China}}} 

\affil[5]{\orgname{Peng Cheng Laboratory}, \orgaddress{\state{Shenzhen}, \country{China}}}

\affil[6]{\orgdiv{School of Computer Science},  \orgname{Peking University}, \orgaddress{\state{Beijing}, \country{China}}}

\affil[7]{\orgdiv{School of Electronic and Computer Engineering},  \orgname{Peking University}, \orgaddress{\state{Shenzhen}, \country{China}}}

\abstract{
Combining Color and Event cameras (also called Dynamic Vision Sensors, DVS) for robust object tracking is a newly emerging research topic in recent years. Existing color-event tracking frameworks usually contain multiple scattered modules which may lead to low efficiency and high computational complexity, including feature extraction, fusion, matching, interactive learning, etc. In this paper, we propose a single-stage backbone network for Color-Event Unified Tracking (CEUTrack) that achieves the above functions simultaneously. Given the event points and color frames, we first transform the points into voxels and crop the template and search regions for both modalities, respectively. Then, these regions are projected into tokens and jointly fed into the adaptive vision transformer network. The output features will be fed into a tracking head for target object localization.  Our proposed CEUTrack is simple, effective, and efficient, achieving over 75 FPS and new SOTA performance. To better validate the effectiveness of our model and address the data deficiency of the color-event tracking task, we propose a generic and large-scale benchmark dataset for color-event tracking, termed COESOT, which contains 90 categories and 1354 video sequences. Furthermore, a new evaluation criterion has been proposed, aiming to better assess tracking results by measuring the difficulty level of video frames. We hope the newly proposed method and dataset provide a better platform for color-event-based tracking. The dataset, toolkit, and source code have been released on  \url{https://github.com/Event-AHU/COESOT}.
}


\maketitle

\section{Introduction}  \label{sec:introduction}

\begin{figure*}[!htp]
\centering
\includegraphics[width=\textwidth]{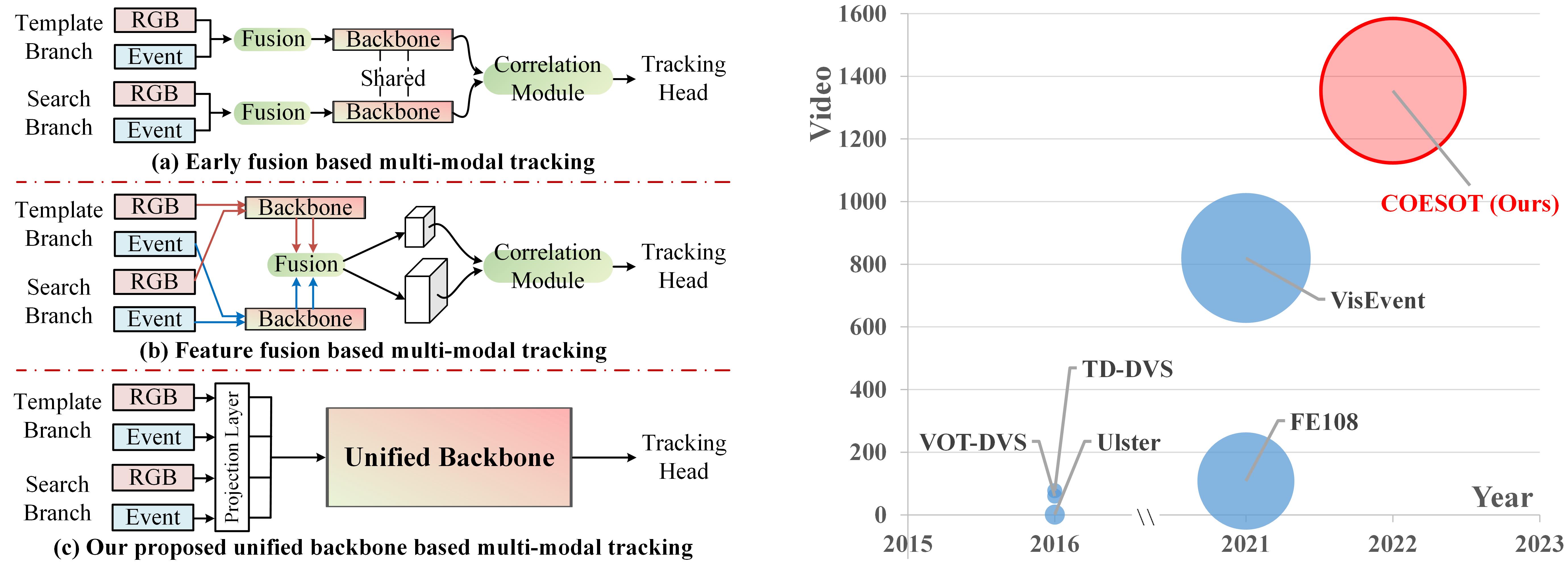}
\caption{
(Left). Comparison of different multi-modal tracking frameworks, including early fusion (EF), middle feature fusion (MF), and our proposed unified tracking framework.
(Right). Comparison of existing color-event tracking datasets. The circle size is proportional to the total frame numbers of the dataset. Best viewed in color.
}
\label{Fig1_comparsion}
\end{figure*}

Visual object tracking targets locating the initialized target object using a series of bounding boxes in the color videos. It is one of the most important tasks in computer vision and provides a good basis for other vision tasks, such as activity recognition, person re-identification, and trajectory analysis. Many representative trackers are proposed like MDNet~\cite{nam2016mdnet}, SiamFC~\cite{siamfc}, SiamRPN~\cite{siamrpn}, DiMP~\cite{bhat2019DiMP}, TransT~\cite{chen2021transt}. 
Although these Color-based trackers exhibit satisfactory outcomes in straightforward scenarios, their efficacy notably falters under intricate challenging scenarios like \emph{low illumination, over-exposure, fast motion, heavy occlusion}, and \emph{background clutter}. 
In these situations, the RGB camera will generate low-quality visible color videos while event camera can promote tracking performance with its bio-inspired sensor and motion-sensitive character.

To address the limitations arising from single-modal defects, some researchers resort to fusing color and event cameras for reliable object tracking. As the event camera is a bio-inspired sensor, which records binary events (i.e., the ON and OFF event) in each pixel when the variation of light intensity exceeds the given threshold. It shows significant advantages over RGB cameras in the \emph{high dynamic range} (HDR), \emph{low energy consumption,} and \emph{low latency}. The dense temporal resolution makes it almost free from the interference of motion blur, and HDR ensures its imaging performance in low-illumination and over-exposure scenarios. Simultaneously, RGB cameras provide color and texture information that hold significant relevance for tracking. The integration of color and event cameras provides a new avenue for practical tracking. 

Despite their scarcity, a handful of studies have delved into the fusion scheme for color-event-based tracking~\cite{wang2023visevent, iccvFE108, huang2018event, gehrig2018asynchronous, zhao2022HNN}.
To be specific, 
Zhang et al.~\cite{iccvFE108} design the cross-domain feature integrator to fuse the color and event data. 
Huang et al.~\cite{huang2018event} reconstruct grey frames from event streams to enhance the feature learning for their SVM-based tracker. 
These algorithms divide the tracking into multiple stages, including \emph{backbone network}, \emph{feature fusion}, \emph{interactive head}, and \emph{tracking head}, as shown in Fig.~\ref{Fig1_comparsion}~(a)-(b). 
Note that, Siamese matching based color-event trackers~\cite{iccvFE108, wang2023visevent} involve the \emph{template} and \emph{search} feature extraction and matching, which further complicates the tracking framework. Further, their models achieve about 20 and 14 FPS only, which makes it hard to achieve real-time tracking in practical applications. This inspired us to think about designing a unified and simple multi-modal tracking framework to achieve efficient and accurate tracking. The recently proposed Transformer~\cite{vaswani2017Transfomer} has shown its natural advantages and effects in simplifying network structure, such as object detection~\cite{carion2020detr}, segmentation~\cite{zheng2021segmentFormer}, and tracking~\cite{swintrack, mixformer}. 
However, there is still no work to design a unified tracking framework for color-event object tracking task.

In this paper, we propose a unified single-stage tracking framework to accomplish color-event tracking, termed CEUTrack. As shown in Fig.~\ref{Fig1_comparsion} (c), it contains three main modules, including the projection embedding layer, unified adaptive backbone, and tracking head. 
With the event stream sampled from DVS, we first reconstruct the event points into voxel sets for efficient representation.
Accompanied by the template and search branch inputs, CEUTrack crops the color frames and event voxel sets to get four regularized pending tensors. 
After projection, these inputs are embedded into multi-modal tokens and fed into the unified vision transformer backbone. To facilitate information flow between different Transformer blocks, we propose the use of adapters to connect various Transformer layers. Specifically, we utilize cross-attention to aggregate the input and output of the Transformers, allowing for better capture of richer multi-level feature representations. The tracking head projects the output tokens to predict the trajectory of the target object. Compared with the popular Siamese frameworks (\textit{dual-branch for unimodal, four-branch for bimodal data}), our proposed CEUTrack involves multi-modal tracking into single-branch architecture that concatenates all of the tokens as one entry. Therefore, it simplifies the multi-modal (color-event) tracking significantly and achieves a very high running efficiency (75 FPS). Extensive experiments demonstrate that our model achieves new SOTA (State-Of-The-Art) performance on previous and proposed color-event tracking datasets.

To better validate the effectiveness of our model and facilitate the community, we also propose a large-scale benchmark dataset for Color-Event multi-modality Single Object Tracking, termed COESOT. It contains 1354 aligned video sequences captured by the DVS346 event camera, in which the training and testing subsets have 827 and 527 videos respectively. The target object of COESOT covers a more comprehensive range of categories than existing color-event tracking datasets~\cite{iccvFE108, wang2023visevent}, like \emph{vehicles, pedestrian, card, toy, bird, monkey, tiger, elephant, zebra, crocodile,} etc. To fully reflect the vital challenging factors in visual and event tracking, 17 attributes are annotated to help detailly evaluate the performance of the trackers. More detailed introductions and comparisons can be found in Section~\ref{COESOTbenchmark} and Table~\ref{benchmarkList}.

In addition, we also propose a new evaluation metric named BreakOut Capability score (BOC) for visual object tracking. Different from existing metrics that treat each video equally to get a precision score, such as PR (Precision Rate), NPR (Normalized Precision Rate), and SR (Success Rate), BOC score focuses more on the prominence of the evaluated tracking algorithm compared with existing ones. In other words, a higher weight proportion will be given for a challenging video, while an ordinary proportion will be assigned for a simple video. Note that the difficulty of each video is comprehensively measured by the accuracy of existing trackers.

To sum up, the main contributions of this paper can be summarized as the following four aspects:

$\bullet$ We propose an adaptive unified tracking framework based on a transformer network, termed CEUTrack, which is the first simplified one-stage backbone for color-event tracking that achieves feature extraction, fusion, and interactive learning simultaneously.

$\bullet$ We propose a large-scale and general benchmark dataset for color-event tracking, termed COESOT. It consists of the most significantly abundant categories of target objects and video sequences in color-event tracking community to date. We extend multiple modern baseline methods on the COESOT for future works' benchmark comparison.

$\bullet$ We propose a new evaluation metric for the tracking task, termed BOC score. It can better reflect the outstanding ability of the tracking algorithm compared with the existing baseline methods on difficult videos.

$\bullet$ Extensive experiments on multiple benchmarks (COESOT, VisEvent, FE108) assess the effectiveness and efficiency of the algorithm. The proposed tracker CEUTrack sets new SOTA performance on existing datasets meanwhile running at a remarkable speed (75 FPS).

\section{Related Work}

In this section, we will briefly review the related works on event-based tracking and RGB-Event based tracking. More works can be found in the \href{https://github.com/wangxiao5791509/VisEvent_SOT_Benchmark/blob/main/Event_Tracking_Paper_List.md}{paper list}.

\begin{figure*}[!htp]
\centering
\includegraphics[width=1\textwidth]{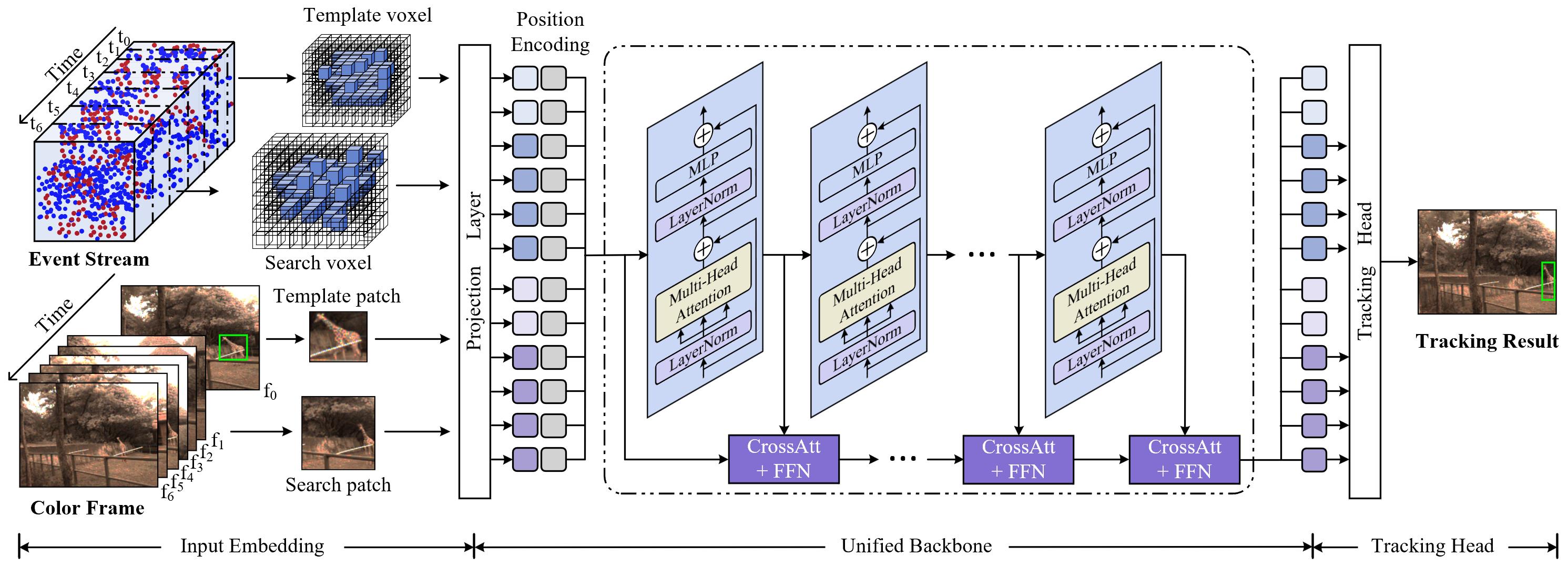}
\caption{The overview of our proposed Color-Event Unified Tracking framework CEUTrack. 
It simplifies the multi-branch multi-modal tracking framework based on the idea of \emph{one branch backbone for all}, which gets rid of cumbersome modules, like multi-stream feature extraction, fusion and correlation, and multi-stage steps.  
} 
\label{framework}
\end{figure*}

\noindent
\textbf{Event-based Tracking.} 
Event-based tracking has attracted more and more attention in recent years. To be specific, based on the DVS camera, ATSLTD~\cite{chen2019asynchronous} proposes an event-frame combined algorithm to construct the event to frames and warp the spatio-temporal information of asynchronous retinal events. 
Chen et al.~\cite{chen2020end} further feed the sequence of TSLTD frames to a retinal motion regression network to perform an end-to-end 5-degree-of-freedom (5-DoF) object motion regression.  
Ramesh et al.~\cite{ramesh2018long} use the discriminative representation for the object with online learning for long-term event-based tracking. 
Chamorro et al.~\cite{chamorro2020high} present an ultra-fast tracking algorithm able to estimate 6-DoF motion and a robust mechanism for the matching of events with projected line segments with very fast outlier rejection.  
Ignacio et al.~\cite{alzugaray2020haste} propose an asynchronous patch-feature tracker that relies solely on events and processes each event individually as soon as it gets generated. 
To solve the HDR situation, Jiang et al.~\cite{jiang2020object} combined the offline-trained detector with an online-trained tracker in a specific Kalman fusion scheme. 
SiamEvent~\cite{chae2021siamevent} correlates the embedded events at two timestamps to compute the target edge similarity and use the Siamese network via edge-aware similarity learning. 
Zhu et al.~\cite{zhu2022NIPSeventTrack} proposes an event-based tracker based on key-event embedding and motion-aware target likelihood prediction. 
STNet~\cite{zhang2022stnet} employs the Transformer network and spiking neural network (SNN) to extract temporal and global spatial information. 
However, with the above development, event-based tracking still has low tracking ability because of the limited event-modality representation and lack of color information. 

\noindent
\textbf{Color-Event based Tracking. }
Tracking by combining color frames and event streams is an interesting and reliable way to achieve high-performance target object localization. For instance, Daniel et al.~\cite{gehrig2018asynchronous, gehrig2020eklt} firstly leverage the complementarity of event and visible cameras to realize low latency feature tracking. DashNet~\cite{yang2019dashnet} jointly processes the synchronous activation from artificial neural networks and SNN spikes signal resources based on the complementary filter and attention module. ESVM~\cite{huang2018event} is a color-event tracker proposed by Huang et al., which targets tracking objects by adaptive search region mining and online SVM (Support Vector Machine) updating. Liu et al.~\cite{liu2016combined} propose a three-step tracking algorithm by fusing color frames and event streams simultaneously. Zhang et al.~\cite{iccvFE108, zhang2023universal} propose self- and cross-domain attention with an adaptive weighting scheme to fuse visual and event images. Wang et al.~\cite{wang2023visevent} build the CMT which is a cross-modality transformer module that performs well for color-event tracking. 
Zhang et al. propose the AFNet~\cite{zhang2023frame} which is a dual-branch network with modality-alignment modules to export the information fusion of color and event frames to realize high-frame rate tracking. 
Although good performance can be achieved in some scenarios, these trackers involve multiple components and branches, such as fusion strategies, multi-step structures, augmentation modules, and dual-branch (Siamese) frameworks to locate the target object. Therefore, these algorithms suffer from heavily limited tracking efficiency with superfluous module design. In this paper, we propose a simple and unified color-event tracking framework that can realize more efficient and accurate tracking.

\section{Methodology}

\subsection{Overview} 
As aforementioned in the previous paragraph, most existing multi-modal visual trackers are developed based on classification, correlation filters, or Siamese tracking frameworks. Researchers focus on designing new fusion modules for bimodal-based tracking tasks, like the color-event data studied in this work. Therefore, the currently obtained frameworks may be too complicated and slow to employ in practical applications. In this work, we propose a simplified color-event tracking framework, termed CEUTrack, as shown in Fig.~\ref{framework}. Given the event streams, we first transform them into voxel representation via voxelization operation. Then, we crop the template patch and voxel according to the initialized location from the color frame and event voxel respectively. The search patch and voxel are also extracted from search regions. A project layer is adopted to transform the four inputs into token representations. The input embeddings and position encoding features are concatenated as a unified representation of our tracker. Then, we propose a unified Transformer network as the backbone to learn the spatial-temporal feature representation. Finally, the obtained tokens are fed into the tracking head for target object localization. More details about each module will be introduced in the following paragraphs respectively.


\subsection{Input Representation}
In this paper, we denote the color frames as $\mathcal{I}_r \in \mathbb{R}^{W \times H \times 3}$ and the event streams as $\mathcal{E}_s$,  
where $r \in \{1, 2, ... , N\}$ is the index of frame, $s \in \{1, 2, ... , S\}$ is the index of event point. $H, W$ are the height and width of video frames. Usually, each event point $e_s$ in $\mathcal{E}_s$ is formulated as a quadruple form $\left[x_{s}, y_{s}, t_{s}, p_{s}\right]$. $(x_s, y_s)$ are pixel location coordinates of the event point $e_s$; $t_s$ is the timestamp, and $p_s \in \{-1, 1\}$ is the binary polarity, i.e., the positive or negative event which corresponds to the blue or red point in Fig.~\ref{framework}. To make full use of the CNN network, previous researchers usually transform the asynchronous event streams into image-like representations for visual tracking. However, the transition from asynchronous to synchronous will disrupt bulk temporal information.

In this paper, we transform the set of event points into a voxel to preserve the temporal information well, and exploit sparse representation of event data to reduce the computational complexity via the top-$k$ voxel selection mechanism. 
Specifically, we first collect the event stream shards $\{e_s\}^r_{r-1}$ in the time interval of two adjacent color frames $r-1$ and $r$.  
Inspired by the typical setting of 3D perception models~\cite{yan2018second, shi2020pv, yin2021center}, $\{e_s\}^r_{r-1}$ is transformed into spaced voxels $ \mathcal{V}_s = \{ [V_x, V_y, V_z, V_f] \}^{r}_{r-1} \in \mathbb{R}^{Q\times L}$. Here, $V_x, V_y, V_z$ denotes the three-dimensional coordinate in the 3D vision by treating the temporal view as the $z$ coordinate. $V_f$ is the feature representation of voxel $\mathcal{V}_s$. The dimension of $L$ is 19, which consists of three coordinate values and voxel features with length $1\times 16$.  
For the event with range $ W\times H\times T $ along the $ x\times y\times z $ respectively, the size of voxels is $ v_W \times v_H \times v_T $ accordingly. Following the 3D detection benchmark SECOND~\cite{yan2018second}, the voxel location is accessed by the center of grid coordinates. The total number of voxels $ Q = {\frac{W}{v_W} \times \frac{H}{v_H} \times \frac{T}{v_T}}$.  
By recapitulating the dense event points into sparse voxel grids, the computation can be reduced significantly in the training and inference phase. 
%
For our framework, given the color frame and event voxel grids set, we first crop the template and search region which is 2 and 4 times larger than the initialized target location. Then, we resize the template and search patches for the color frame modality into $\textbf{z}^f \in \mathbb{R}^{H_z \times W_z \times 3}$ and $\textbf{x}^f \in \mathbb{R}^{H_x \times W_x \times 3}$. 
For the event modality, we remove the voxel grids whose coordinates are not in the search and template region.
Then, we sort the voxel grids based on the density of events and select the top $M_x=4096$ and $M_z=1024$ grids to sparse the input dimension. Therefore, we can get the template voxels $\textbf{z}^{v} \in \mathbb{R}^{M_z \times L}$ and search voxels $\textbf{x}^{v} \in \mathbb{R}^{M_x \times L}$. Finally, we build a to-be-input appearance tuple in each frame as $(\textbf{z}^f, \textbf{x}^f, \textbf{z}^{v}, \textbf{x}^{v})$ for the proposed unified framework, as shown in Fig.~\ref{framework}.

\subsection{Unified Backbone Network} 
Before feeding the inputs into Transformer network, we introduce a projection layer to transform them into token sequence representations. The projection layer consists of four parallel and split Conv$\_$BN$\_$ReLU blocks. 
For the color frame, the corresponding block project the template $\textbf{z}^f$ and search patches  $\textbf{x}^f$  into feature embeddings $F^f_z \in \mathbb{R}^{N_z \times C}$ and $F^f_x\in \mathbb{R}^{N_x \times C}$ using two non-shared $16 \times 16$ convolutional (Conv) operators. Note that, the $N_z$ and $N_x$ are the length of template and search token sequences. 
For the event voxel, we project the search $\textbf{x}^{v}$ and template voxel $\textbf{z}^{v}$ into $F^v_x \in \mathbb{R}^{N_x \times C}$ and $F^v_z \in \mathbb{R}^{N_z \times C}$ based on two non-shared $4\times4$ convolutional operators. In addition, we take the position embedding into consideration and add them with feature embeddings. More in detail, the position embedding is shared between search regions of the color frame and event voxel domain, i.e., $P_x \in \mathbb{R}^{N_x \times 1}$. Similarly, shared position embedding can be obtained for template regions, i.e., $P_z \in \mathbb{R}^{N_z \times 1}$.  


With the projected token tuple ($F^v_z, F^v_x, F^f_z, F^f_x$) obtained, we merge them into a unified token $\mathbf{U}_s \in \mathbb{R}^{2(N_z+N_x)\times C}$ along the dimension of sequence length, and feed into the stacked and cascade twelve Transformer layers. 
As shown in Fig.~\ref{framework}, 
each backbone layer consists of a standard Transformer block and an adapter block. 
Each Transformer block consists of two layernorm (LN), multi-head self-attention (MSA), and multi-layer perception (MLP) with residual connections. 
The detailed computation process can be formulated as: 
\begin{equation}
\small 
\begin{split}
& \text{MSA}(\mathbf{Q}, \mathbf{K}, \mathbf{V})  = Softmax({\frac{\mathbf{Q}\mathbf{K}^{\mathsf{T}}}{\sqrt{d_{k}}}}) \cdot \mathbf{V},\\
& \tilde{\mathbf{U}}_s  = \mathbf{U}_s + \text{MSA}(LN (\mathbf{U}_s, \mathbf{U}_s, \mathbf{U}_s)), \\
& \mathbf{U}_s  = \tilde{\mathbf{U}}_s  + \text{MLP}(LN (\tilde{\mathbf{U}}_s)). 
\end{split}
\end{equation}
We adopt the vanilla ViT-B~\cite{dosovitskiy2020image} to build the Transformer blocks, which provide a clean architecture and publicly available pre-trained weights. 
This adaptive backbone avoids the tedious process including modality interactive learning, fusion, and matching operations between two branches which are popular in existing multi-modal visual tracking algorithms. 
Further, instead of employing ViT directly, we introduce an adapter block for each nearby ViT block. As shown in Fig.~\ref{framework}, the adapter module consists of twelve blocks and each block is only built by a cross-attention and FFN layer with residual connection. 
The adapter attention shares the learnable position embedding from the backbone. 
With the simple fusion way, we merge and interact the feature between nearby layers and search-template frames in one step. 

To obtain the location of target object, we employ a popular tracking head to directly estimate the center position and scale of the bounding box. Specifically, the tracking head contains three branches that predict the classification score, center offset, and the size of bounding box, respectively. Each branch consists of four $3 \times 3$~Conv$\_$BN$\_$ReLU  and a $1 \times 1$~Conv  layer respectively.

\begin{table*}[!thb]
\centering
\footnotesize
\caption{Frame and event camera based datasets for single object tracking.} \label{benchmarkList}
\resizebox{\textwidth}{!}{ 
\begin{tabular}{l|cccccccccccccccc}
\hline 
\textbf{Datasets}    &\textbf{Year}	&\textbf{\#Videos}  &\textbf{\#Frames} &\textbf{\#Class}    &\textbf{\#Att} &\textbf{\#Resolution} &\textbf{Aim}   &\textbf{Absent}  &\textbf{Color} &\textbf{Real}  &\textbf{Public}  \\ 
\hline
\textbf{VOT-DVS}~\cite{hu2016dvs}     &2016    &60           &-      &-    	 &-    &$240 \times 180$  &Eval  &\xmark   &\xmark     &\xmark     &\cmark     \\
\textbf{TD-DVS}~\cite{hu2016dvs}        &2016     &77          &-      &-    	 &-  &$240 \times 180$  &Eval  &\xmark    &\xmark     &\xmark   &\cmark     \\
\textbf{Ulster}~\cite{liu2016combined}   &2016      &1     &9,000  		   &-    	 &-    &$240 \times 180$ 
 &Eval   &\xmark  &\xmark     &\cmark     &\xmark    		\\
\textbf{EED}~\cite{mitrokhin2018event} &2018     &7     &234   &  -  	 &-   &$240 \times 180$   &Eval  &\xmark   &\xmark      &\cmark     &\cmark     \\
\textbf{FE108}~\cite{iccvFE108}	&2021		&$108$     &208,672  	& 21  & 4     &$346 \times 260$     &Train \& Eval   &\xmark     &\xmark     &\cmark       &\cmark    \\
\textbf{VisEvent}~\cite{wang2023visevent}  	&2021     &$820$       &371,127      	&  -  	&17     &$346 \times 260$  &Train \& Eval      &\cmark &\cmark          &\cmark           &\cmark     \\
\hline 
\textbf{COESOT}  	&2023     &\textcolor{black}{\textbf{1354}}      &\textcolor{black}{\textbf{478,721}}       &\textcolor{black}{\textbf{90}}    	&17   &$346 \times 260$    & Train \& Eval  &\cmark &\cmark  &\cmark   &\cmark     \\
\hline 
\end{tabular}
}
\end{table*}

\subsection{Training and Testing Phase} 

To train our proposed CEUTrack effectively, we combine three loss functions, including focal loss~\cite{lin2017focal} for classification, L1 loss and GIOU loss~\cite{rezatofighi2019generalized} for bounding box regression. The overall loss function can be written as: 
\begin{equation}
    L = \lambda_1 L_{focal}(y,y^{\prime}) + \lambda_2 L_{L1}(b, \hat{b}) + \lambda_3  L_{giou}(b, \hat{b})
\end{equation} 
where $(y$, $y^{\prime})$ separately are the ground truth classification label and predicted class, $(b, \hat{b})$ shows the ground-truth box and prediction box coordinate.
The trade-off parameters $\lambda_1$, $\lambda_2$, and $\lambda_3$ are set as 1, 1, and 14, respectively. Weight parameters are manually set to balance the magnitude of loss functions without hyperparametric searching.

In our inference procedure, we first crop the template color patch from the first frame based on the initialized bounding box. The template color patch is resized into a fixed scale, i.e., $128\times 128\times 3$, in our experiment. For the event voxel, we get its template representation via the following two rules: 
1) the coordinate of the voxel grid is within the cropped template region; 
2) top-1024 voxel grids are computed based on the density of event points. For the template with voxels less than 1024, zero-value grids will be used for padding. The pre-processed template voxel patch and color patch will be saved into the cache for subsequent tracking. 
For the following tracking, we crop the search color patch for the color domain and resize it into $256\times 256\times 3$. We select top-4096 voxels by following the aforementioned rules as the search voxel patch. Then, we combine the search and template patches of both modalities for tracking by feeding them into the unified network and head for target object localization. The same operations are executed for the subsequent color frames and event voxels until the end of the testing operation.

\section{COESOT Benchmark Dataset} \label{COESOTbenchmark}

\subsection{Hardware for COESOT Collection} 
The video sequences in COESOT are all collected by the color and event camera DVS346 with a zoom lens. Different from previous datasets FE108~\cite{iccvFE108} and VisEvent~\cite{wang2023visevent}, which collect sequences without any distance and focal length change, COESOT collects videos with many scale-changing scenarios employing zoom lens cameras. As shown in Fig.~\ref{Fig1_comparsion} (right sub-figure), our COESOT contains more than 1300 video sequences, which is the most large-scale and modal-aligned well dataset to date for the current color-event community and outperforms existing datasets with a significant margin. Some sequences of proposed COESOT are visualized in Fig.~\ref{dataset_vis}. The event image represents a well-motion situation when the color frames are reported not very well.


\subsection{Data Collection~} 
\begin{figure*}[!t]
\center
\includegraphics[width=\textwidth]{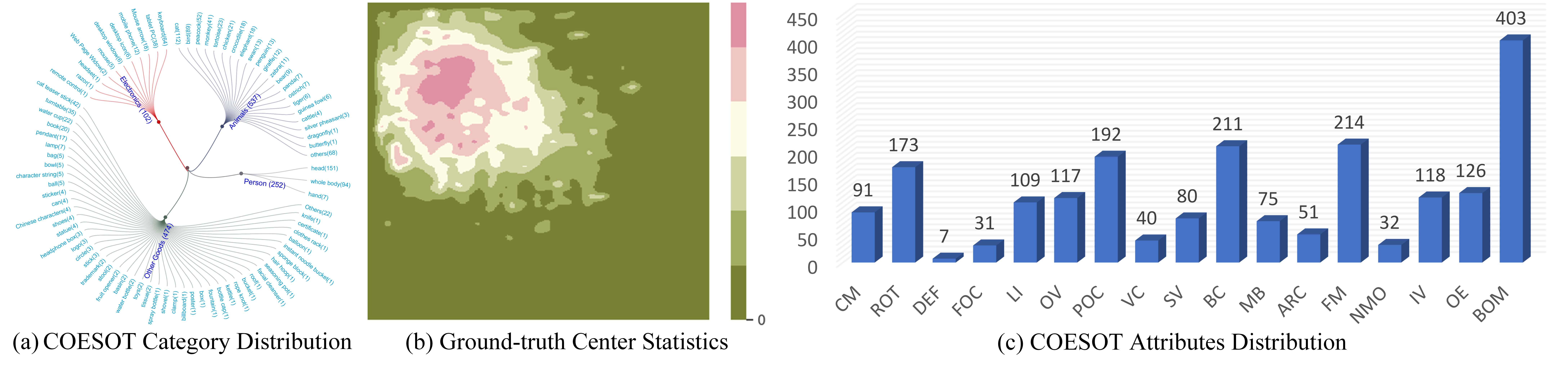}
\caption{Details of attributes, category, and distribution of COESOT dataset.} 
\label{COESOT_attributes} 
\end{figure*}

\begin{figure*}[!t]
\centering
\includegraphics[width=\textwidth]{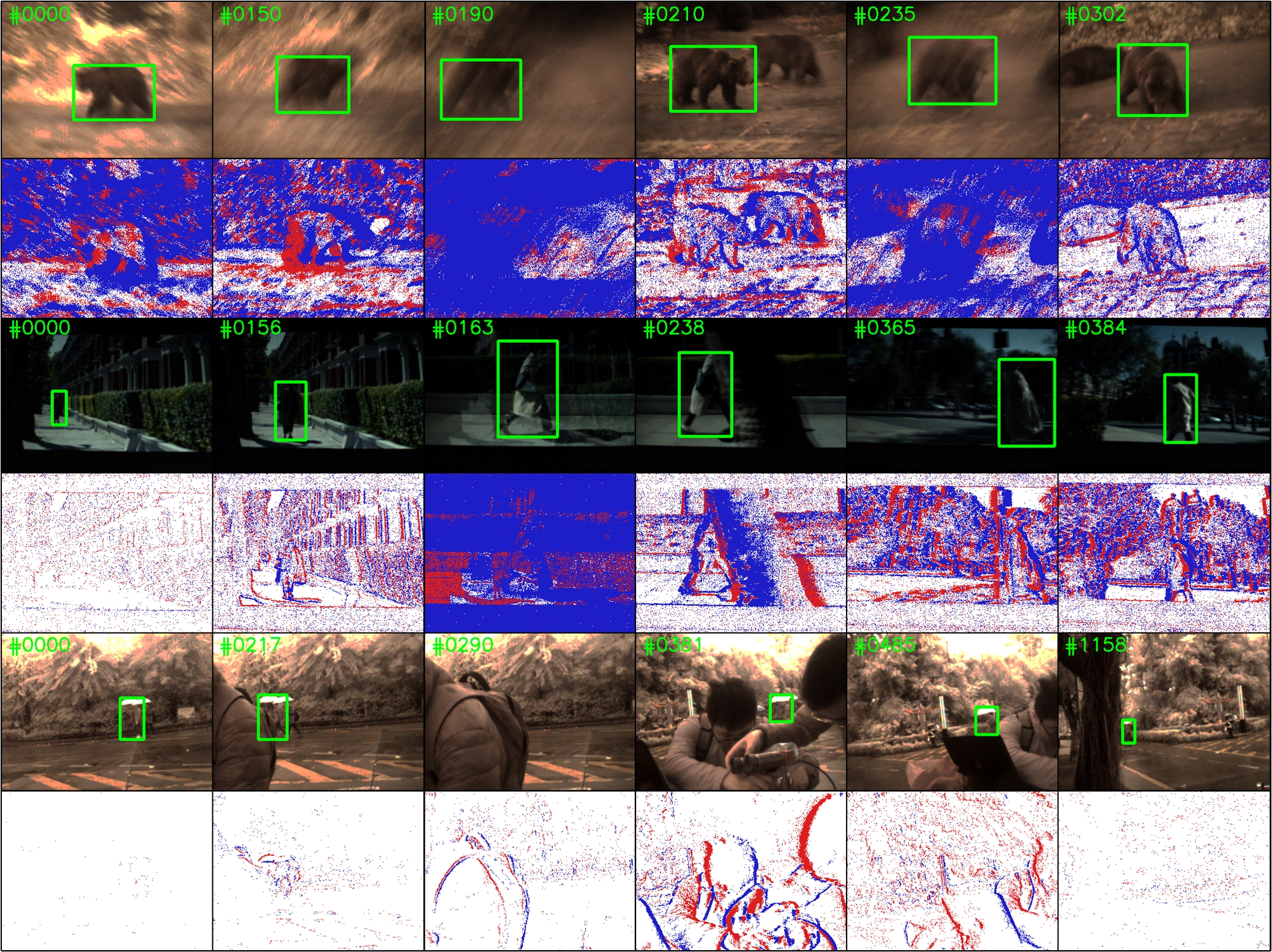}
\caption{Some representative examples from our proposed COESOT test set. 
 } 
\label{dataset_vis}
\end{figure*}

Current existing color-event tracking datasets are all small-scale~\cite{hu2016dvs, liu2016combined, mitrokhin2018event}, frames lack color information~\cite{iccvFE108}, or timestamp misaligned of partial data~\cite{wang2023visevent}. Further, the category of the target object is limited, as illustrated in Table~\ref{benchmarkList}. 
In this work, we propose a general dataset for \textbf{Co}lor-\textbf{E}vent camera based  \textbf{S}ingle \textbf{O}bject \textbf{T}racking, termed COESOT. It contains 1354 color-event videos with 478,721 color frames and corresponding event streams. We split sequences into a training and testing subset, which contains 827 and 527 videos, respectively. The videos are collected from both outdoor and indoor scenarios (e.g., \textit{street}, \textit{zoo}, and \textit{home}) using the DVS346 event camera with a zoom lens. Therefore, our videos can reflect the variation in the distance at depth, but other datasets are failed to. Different from existing benchmarks which contain limited categories, our proposed COESOT covers a wider range of object categories (90 classes), as shown in Fig.~\ref{COESOT_attributes} (a).

\subsection{Video Annotation~} 
The ground truth of the proposed COESOT dataset is densely annotated, i.e., in a frame-by-frame way. The absent flag of each frame is also labelled to help researchers design their trackers. 
Inspired by VisEvent~\cite{wang2023visevent}, we annotate each testing video sequence with 17 attributes to evaluate trackers in specific challenging environments, e.g., full occlusion (FOC), viewpoint Change (VC), rotation (ROT), fast motion (FM), partially occlusion (POC), low illumination (LI), scale variation (SV), background object motion (BOM), motion blur (MB), over-exposure (OE), etc. The statistical distribution of the ground truth center position is shown in Fig.~\ref{COESOT_attributes} (b). The distribution of videos in each attribute is shown in Fig.~\ref{COESOT_attributes} (c).

\subsection{Evaluation Metrics~} 
For the evaluation measurement of COESOT,  we follow the popular metrics success rate (\textbf{SR}), precision rate (\textbf{PR}), normalized precision rate (\textbf{NPR}), and our newly proposed \textbf{BOC}.
SR is computed as the ratio of the number of successfully tracked frames. PR is calculated by comparing the distance between the centers of the ground truth bounding box and the tracking result in pixels. Since PR is sensitive to target size and image resolution, NPR normalizes the precision as in ~\cite{trackingnet} to ensure the consistency of evaluation across different target scales.

For the motivation of BOC metric, existing evaluation metrics treat each video in a dataset equally and report an overall result. We perceive that the difficulty of each video is different, and the newly proposed high-quality visual tracker should be given a higher reward if it succeeds in tracking an object while previous methods failed. With this in mind, we first compute the difficulty of each video in the COESOT dataset based on the SR score of the tested baseline trackers. Then, we weigh the tracking result of each video of the evaluated tracker using the obtained difficulty value. Thus, our proposed BOC (BreakOut Capability) score effectively validates the real effectiveness and conspicuousness compared with existing trackers. 
If there are $N$ testing videos and $T$ compared base trackers, the formulation of BOC can be written as: 
\begin{equation}
    \label{BOCscore} 
    \small 
     \textbf{BOC}(evalT) = \frac{1}{N}\sum_{i=1}^{N}[ \text{SR}(evalT^i) * 
     (1 - \frac{1}{T} \sum_{t=1}^{T} \text{SR}(baseT_t^i))], 
\end{equation}
\textcolor{black}{where $\text{SR}(evalT^i)$ denotes the success rate score of the evaluated tracker on video $i$, 
$\text{SR}(baseT_t^i)$ denotes the SR score of compared base tracker $t$ on the video $i$. $(1 - \frac{1}{T} \sum_{t=1}^{T} \text{SR}(baseT_t^i))$ indicates the difficulty degree of video $i$ based on tracking results of baseline trackers, which is employed as the weight coefficient of video $i$.}

\subsection{Benchmarked Trackers}

With the single-modality input of the benchmarked baseline trackers, we feed the 25 trackers with an early-fusion (EF) strategy. Specifically, we stack the event stream into images and fuse the generated event images with the color images at the rate of 0.2:1. The baseline trackers assessed in our benchmark can be divided into three groups:

- \textbf{Four Siamese-based trackers}: SiamRCNN~\cite{siamr-cnn}, TrSiam~\cite{trdimp}, SiamRPN~\cite{siamrpn}, and  SiamFC~\cite{siamfc}.

- \textbf{Eight Transformer-based trackers}: AiATrack~\cite{gao2022aiatrack}, OSTrack~\cite{ye2022OSTrack}, MixFormer22k~\cite{mixformer},  MixFormer1k~\cite{mixformer},  TransT~\cite{chen2021transt}, STARK-ST101~\cite{stark}, STARK-ST50~\cite{stark}, and STARK-S50~\cite{stark}.

- \textbf{Thirteen Discriminant-based trackers}: RTS~\cite{paul2022rts}, ToMP101~\cite{tomp}, ToMP50~\cite{tomp}, TrDimp~\cite{trdimp}, KeepTrack~\cite{keeptrack}, KYS~\cite{kys}, PrDiMP50~\cite{prdimp}, PrDiMP18~\cite{prdimp}, SuperDiMP~\cite{superdimp}, SuperDiMP$_{\text{Simple}}$~\cite{superdimp}, DiMP18~\cite{dimp}, DiMP50~\cite{dimp}, and ATOM~\cite{atom}.

Additionally, three algorithms that follow the middle-level feature fusion (MF) strategy are re-trained and evaluated, including MDNet-MF~\cite{nam2016mdnet}, VITAL-MF~\cite{song2018vital}, and SiamFC-MF~\cite{siamfc}.

\subsection{Comparison with Existing Datasets}

\noindent
\textbf{COESOT \textit{vs.} FE108 Dataset.} 
This is a tracking dataset collected by a grey-scale DVS346 camera. Because of the characteristic of the camera, the visual frames captured are grey-scale and the timestamps of events are all aligned with gray frames. 
Note that the videos in the FE108 dataset are all collected in indoor scenarios without the zoom lens. FE108 contains 108 videos in 21 object target categories, which consist of three groups: animals, vehicles, and daily goods. It annotates the target ground-truth with only four attributes: low light, high dynamic range, fast motion with and without motion blur on visual frames. Compared with our proposed COESOT, FE108 contains much fewer target categories, attributes, scenes, sequences, and frames. In addition, our COESOT also provides the absent labels while FE108 is not provided. 

\noindent
\textbf{COESOT \textit{vs.} VisEvent Dataset.} 
VisEvent is a tracking dataset that also collects videos using a camera without a zoom lens. Different from grayscale videos in FE108, VisEvent collected all color videos. Its training and testing subsets contain 500/320 sequences respectively. However, some video sequences in VisEvent are misalignment along time dimension and some sequences lack event source files (\textit{*.aedat4}) which limited the future development on it. In our tracking experiment, the source files of each sequence are essential due to the voxelization transform. Further, the provision of source event files can contribute to community development.
In addition, the synchronization status of color and event modal is necessary for many multi-modal tracking processes. 
Therefore, we select 205 training sequences and 171 testing videos which provide sequence source files and aligned timestamps between two modalities to evaluate the tracker's performance. For a fair comparison, the evaluation performances reported in our paper are all evaluated with the same testing video sequences. 
Compared with our proposed COESOT, VisEvent contains unknown object categories, much fewer frames, and video sequences, especially after filtering useless videos. Therefore, compared with VisEvent, our COESOT provides a high-quality, largest-scale, and most reliable dataset to date.

\section{Experiment} 

\subsection{Implementation Details}   
For FE108 and our proposed COESOT, we trained on the training subsets and evaluated on testing subsets,  respectively. For VisEvent, we select 205 and 172 videos with source aedat4 files for training and testing. Without any specific fine-tuning operation, the three datasets are trained with the same settings. 
In the training phase, we randomly sample the template and search images in a 100-frame window. The model is trained with an AdamW optimizer with a weight decay of 1e-4 and a batch size of 32. The initial learning rate is set at 1e-4 and decays in magnitude at the 40th epoch. The training epochs are set at 50, with 60k frame pairs sampled every epoch. The dropout rates of transformer layers in CEUTrack are all set to zero. Note that, unlike currently popular training strategies, our training process does not employ any data augmentation strategy (horizontal flip, brightness jitter, image blur, relative shift, etc.) Our source code is implemented on Python 3.8 and Pytorch 1.8 with an RTX-3090 GPU.

\subsection{Comparison with Other SOTA Trackers} 
In this subsection, three frame-event datasets are evaluated in our experiments, including FE108, VisEvent, and the newly proposed COESOT.

\begin{figure}
    \centering
    \includegraphics[width=1\columnwidth]{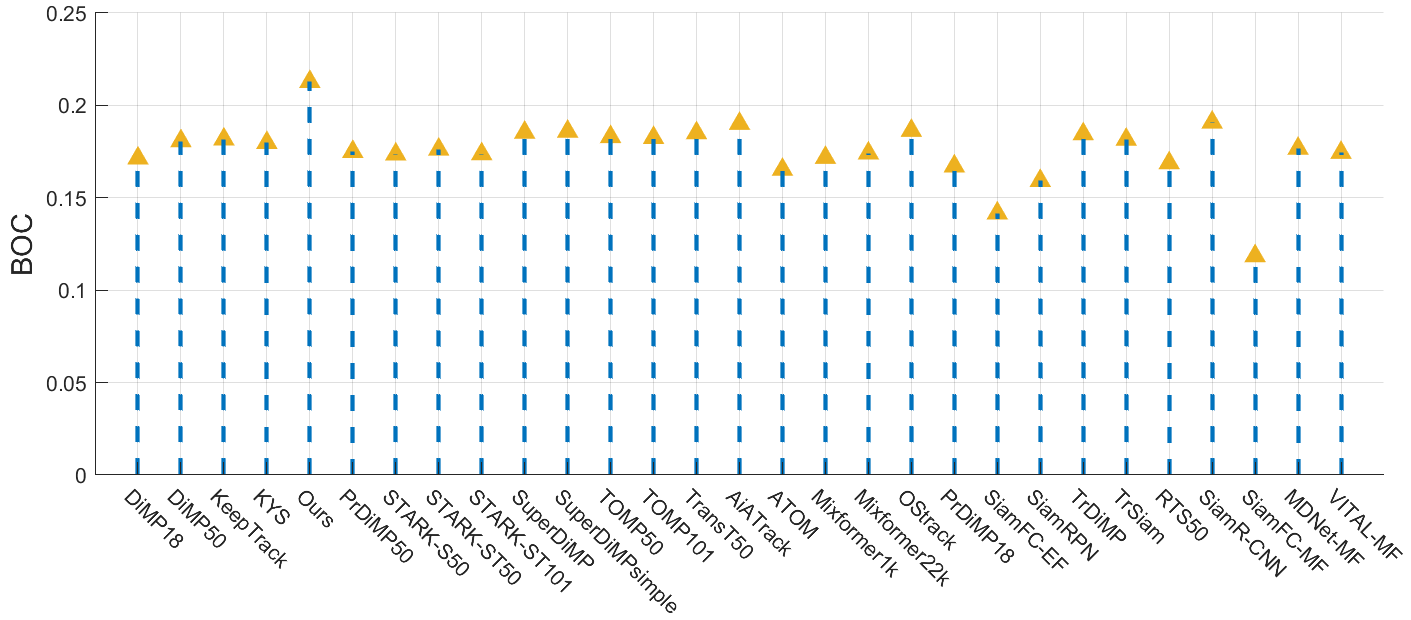}
    \caption{BOC scores comparison of baseline trackers on the COESOT dataset.} 
    \label{BOCscore_fig}
\end{figure}

\noindent
\textbf{Results on COESOT.~}As depicted in Table~\ref{COESOT_auc}, we present the results of 28 prominent and recent trackers. Our proposed CEUTrack demonstrates the most remarkable tracking performance, significantly surpassing other strong trackers.
To be specific, we obtained 21.3, 62.7, 76.0, and 74.9, on BOC, SR, PR, and NPR metrics, respectively. In contrast, single-stream Transformer based tracker OSTrack~\cite{ye2022OSTrack} directly employs the fused color-event images as input and gets $18.6, 59.0, 70.7, 70.5$ on these metrics. Compared with OSTrack, our method showcases notable improvement up to $+2.7\%, +3.7\%, +5.3\%, +4.4\%$. Further, we surpass the second-ranking tracker SiamR-CNN~\cite{siamr-cnn} on the COESOT benchmark by a margin of  $+1.2\%, +1.8\%, +5.0\%, +4.4\%$. 
Our work demonstrates that our unified multimodal tracking method is an effective paradigm for Color-Event object tracking. 
From the perspective of BOC score, as shown in Fig.~\ref{BOCscore_fig}, our proposed CEUTrack, SiamR-CNN~\cite{siamr-cnn}, and AiATrack~\cite{gao2022aiatrack} rank the 1st, 2nd, and 3rd, respectively, which fully demonstrates the breakthrough performance of our tracker.

\begin{table}[t]
\centering
\footnotesize
\caption{Overall tracking performance of 28 baseline trackers on COESOT. The trackers are ranked by SR metric. The first, second, and third-place trackers are highlighted in  \textbf{\textcolor{red}{red}}, \textcolor{blue}{blue} and \textcolor{SeaGreen4}{green} colors respectively. } 
 \label{COESOT_auc}
\setlength{\tabcolsep}{1.5mm}{
\begin{tabular}{l|c|cccc|c}
\hline 
\textbf{ Trackers} & \textbf{Source}   & \textbf{SR}  &\textbf{PR}   &\textbf{NPR}   &\textbf{BOC}  &\textbf{FPS}\\
\hline 
\textbf{ CEUTrack (Ours)} & \textbf{--} & \textbf{\textcolor{red}{62.7}}  & \textbf{\textcolor{red}{76.0}}  & \textbf{\textcolor{red}{74.9}}   & \textbf{\textcolor{red}{21.3}}  & \textcolor{SeaGreen4}{75}\\
\textbf{ SiamR-CNN   }  & CVPR20  &  \textcolor{blue}{60.9}   &  71.0 &  70.5      &   \textcolor{blue}{19.1}       &     5     \\
\textbf{ TransT } & CVPR21     &   \textcolor{SeaGreen4}{60.5}      &     \textcolor{blue}{72.4}   &   \textcolor{blue}{71.9}     &  18.5     &   50   \\
\textbf{ SuperDiMP   }   & --        &  60.2       &    72.0    &  \textcolor{SeaGreen4}{71.7}       &  18.5       &  --   \\
\textbf{ TrDiMP   }   &  CVPR21       & 60.1     &     \textcolor{SeaGreen4}{72.2}    & \textcolor{SeaGreen4}{71.7}      & 18.5       &    26      \\
\textbf{ ToMP101   }  &  CVPR22        & 59.9     &    71.6   &  71.3       & 18.3       &   20        \\
\textbf{ ToMP50  }   &  CVPR22       & 59.8          &    70.8  & 70.9      &  18.3    &    25        \\
\textbf{ TrSiam   }   & CVPR21        &  59.7         &   71.4  &  71.0       &   18.2      & 35  \\
\textbf{ KeepTrack   }   &  ICCV21       & 59.6          &  70.9  &  70.7      &  18.2      &   18  \\
\textbf{ OSTrack }   &  ECCV22       & 59.0          &  70.7     &  70.5       & 18.6   &  \textcolor{blue}{105}   \\
\textbf{ AiATrack   }   &  ECCV22       &  59.0        &   \textcolor{blue}{72.4}    & 71.4         &   \textcolor{SeaGreen4}{19.0}        &   38 \\
\textbf{ DiMP50   }  &  ICCV19        &  58.9         &  72.0  &   71.6     &    18.1         &   43     \\
\textbf{ KYS   }   &   ECCV20      &  58.6        &  71.6  &  71.3       & 18.00     &   20       \\
\textbf{ PrDiMP50   }   & CVPR20        & 57.9         & 69.6   & 69.4    & 17.5      &   30      \\
\textbf{ PrDiMP18   }  &  CVPR20        &  56.7        & 68.0   & 68.4     & 17.7  &   40      \\
\textbf{ DiMP18   }   &   ICCV19      &  56.7        &   69.1  &  69.2      & 17.1  &   57      \\
\textbf{ MDNet-MF   }   & CVPR16        &  56.3    &   69.0  &  70.0         & 17.7     &  14    \\
\textbf{ RTS50   }  &  ECCV22        &  56.1        & 65.1    &  63.2    &  16.9    &   30      \\
\textbf{ STARK-ST50   }   & ICCV21        &  56.0        &  67.7  &  67.2    &  17.6     &  42      \\
\textbf{ MixFormer1k   }   &  CVPR22       & 56.0         &  66.6 &  66.3    &  17.2  &  25       \\
\textbf{ STARK-ST101   }   &  ICCV21       & 55.8         &  67.1 &  67.0    &  17.4     & 32       \\
\textbf{ STARK-S50   }   &   ICCV21      &   55.7      &  66.7   &  66.5     &  17.4  &  42        \\
\textbf{ MixFormer22k   }   & CVPR22        &  55.7        &   66.3 &  66.3  &  17.4    & 25  \\
\textbf{ VITAL-MF  }   & CVPR18    &  55.6   & 68.5   & 69.3   &  17.4  &   16   \\
\textbf{ ATOM   }   &  CVPR19       &  55.0    &   68.8  &  68.3     &  16.5    &     30  \\
\textbf{ SiamRPN   }   &  CVPR18       &   53.5        &  65.7 &  66.2   &  15.9  & \textcolor{red}{\textbf{160}}  \\
\textbf{ SiamFC-EF   }   &  ECCVW16       &   48.4       &   58.3 & 60.5   &  14.2    &  58    \\
\textbf{ SiamFC-MF   }   & ECCVW16      &  41.8   & 49.4 &  50.0   &  11.9   &  58    \\
\hline 
\end{tabular}} 
\end{table}

\begin{table}
\centering
\caption{Experimental results on VisEvent dataset. The best results are highlighted in \textbf{bold}.} 
\label{Viseventtable}
\setlength{\tabcolsep}{0.9mm}{
\begin{tabular}{l|cccccc}
\hline
\textbf{Trackers}  & \textbf{AUC}       &\textbf{SR$_{0.5}$}   &\textbf{SR$_{0.75}$}   &\textbf{PR} &\textbf{NPR} \\   
\hline
\textbf{CEUTrack (Ours)} & \textbf{53.12} & \textbf{64.89}   & \textbf{45.82} & \textbf{69.06} & \textbf{73.81} \\
\textbf{LTMU (EF)} & 49.30 &  60.10   & 37.05 & 66.76 & 69.78\\
\textbf{PrDiMP (EF)} & 48.34 &  57.20   & 37.39 & 64.47 & 67.02\\
\textbf{CMT-MDNet} & 47.51 & 57.44   & 31.22 & 67.20  & 69.78\\
\textbf{ATOM (EF) }  & 44.74  & 53.26 & 31.34  & 60.45 & 63.41\\
\textbf{AFNet (EF) } &44.5 &-  &-  & 59.3 &-\\  

\textbf{SiamRPN++ (EF) } & 44.68  & 54.11 & 33.66  & 60.58 & 64.72\\
\textbf{SiamCAR (EF) } & 43.51 & 52.66 & 34.49 & 58.86  & 62.99\\
\textbf{Ocean (EF) }  & 37.51  & 43.56 & 23.26 & 52.02 & 54.21\\
\textbf{SuperDiMP (EF) } & 33.47 & 36.21   & 17.84 & 46.99 &42.84 \\
\hline   
\textbf{STNet (Event-Only)} & 35.5 & 39.7  & 20.4 & 49.2 &- \\
\textbf{TransT (Event-Only)} & 32.9 & 39.5  & 18.0 & 47.1  &- \\
\textbf{STARK (Event-Only)} & 32.7 & 34.8  & 21.4 & 41.8  &- \\
\hline
\end{tabular}
} 
\end{table}

\begin{table*}[th]
\centering
\caption{Experimental results on FE108 dataset. The best results are shown in \textbf{bold}. }
\resizebox{\textwidth}{!}{
\begin{tabular}{c|lllllllllll}
\toprule
Tracker& \multicolumn{1}{c}{SiamRPN}  & \multicolumn{1}{c}{SiamBAN} & \multicolumn{1}{c}{SiamFC++}  
& \multicolumn{1}{c}{KYS}  & \multicolumn{1}{c}{CLNet}  & \multicolumn{1}{c}{CMT-MDNet}  & \multicolumn{1}{c}{ATOM} 
& \multicolumn{1}{c}{DiMP}  & \multicolumn{1}{c}{PrDiMP}   & \multicolumn{1}{c}{CMT-ATOM} & \multicolumn{1}{c}{\textbf{CEUTrack (Ours)}}  \\
\midrule
SR  & \multicolumn{1}{c}{21.8} & \multicolumn{1}{c}{22.5}   & \multicolumn{1}{c}{23.8}   & \multicolumn{1}{c}{26.6}  & \multicolumn{1}{c}{34.4} &\multicolumn{1}{c}{35.1}  & \multicolumn{1}{c}{46.5} &\multicolumn{1}{c}{52.6} &\multicolumn{1}{c}{53.0} &\multicolumn{1}{c}{54.3}  & \multicolumn{1}{c}{\textbf{55.58}}\\
PR  & \multicolumn{1}{c}{33.5} & \multicolumn{1}{c}{37.4}   & \multicolumn{1}{c}{39.1}   & \multicolumn{1}{c}{41.0}  & \multicolumn{1}{c}{55.5}  & \multicolumn{1}{c}{57.8}  &\multicolumn{1}{c}{71.3}  &\multicolumn{1}{c}{79.1}  &\multicolumn{1}{c}{80.5} 
 &\multicolumn{1}{c}{79.4} & \multicolumn{1}{c}{\textbf{84.46}}\\
\bottomrule
\end{tabular}
} 
\label{fe108table}
\end{table*}

\noindent
\textbf{Results on VisEvent.} 
As shown in Table~\ref{Viseventtable}, our proposed CEUTrack outperforms other SOTA trackers significantly, including PrDiMP~\cite{prdimp}, STNet~\cite{zhang2022stnet}, TransT~\cite{chen2021transt}. 
CEUTrack achieves $53.12, 64.89, 45.82, 69.06, 73.81$ on AUC, SR$_{0.5}$, SR$_{0.75}$, PR, NPR, respectively, which set SOTA performance on this benchmark. Note that, all of the compared methods are evaluated on the same sequences for fair comparison. 
Further, in current Color-Event specifically designed trackers, AFNet~\cite{AFNet} is the newest and most state-of-the-art one. While compared with AFNet, our CEUTrack still owns a significant margin (8.6\% gap in AUC).

\noindent
\textbf{Results on FE108.}
FE108 is a grey-event dataset which has 32 test sequences with grey frames and event stream pairs. As shown in Table~\ref{fe108table}, our model achieves 55.58 and 84.46 on SR and PR metrics, significantly better than the compared trackers (DiMP, PrDiMP, CMT-ATOM, etc.). 
It demonstrates the effectiveness and generalization of our color-event tracker in handling grayscale frames.

\noindent
\textbf{Event-only Comparison on COESOT.}
In our proposed COESOT, we proved both color and event modality data, therefore, besides \textit{color-event tracking}, it can also employed as the benchmark for \textit{event-only tracking}. 
For a complementary and fair comparison for developing future \textit{event-only tracking} methods, we retrain and evaluate 12 baselines on COESOT with event-only as input, as shown in Table~\ref{COESOT_event_only}. We hope COESOT also can be event-only tracking benchmark and contribute to event-based tracking task.

\begin{table}[t]
\centering
\small   
\caption{Event-only tracking performance on COESOT. } 
\label{COESOT_event_only}
\begin{tabular}{l|c|ccccc}
\hline 
\textbf{Trackers} & \textbf{Source}   & \textbf{SR}  &\textbf{PR}   &\textbf{NPR} \\
\hline
\textbf{ TrDiMP }   & CVPR21     &50.7       &56.9       &55.2           \\ 
\textbf{ ToMP50  }   &  CVPR22   &46.3       &52.9       &52.5           \\ 
\textbf{ OSTrack }   &  ECCV22   &50.9       &57.8       &56.7           \\ 
\textbf{ AiATrack   }   &ECCV22   &51.3       &57.9       &56.2           \\ 
\textbf{ STARK   }   &  ICCV21    &40.8       &44.9       &44.4           \\ 
\textbf{ TransT   }   &  CVPR21     &45.6       &51.4       &50.4           \\ 
\textbf{ DiMP50   }  &  ICCV19     &53.8       &61.7       &60.3           \\ 
\textbf{ PrDiMP   }  &  CVPR20     &47.5       &55.1       &54.0           \\ 
\textbf{ KYS   }   &   ECCV20      &42.6       &50.6       &49.7           \\ 
\textbf{ MixFormer   }   & CVPR22   &44.4       &49.4      &48.5           \\ 
\textbf{ ATOM   }   & CVPR19    &42.1       &48.0        &48.1           \\ 
\textbf{ SimTrack   }   & ECCV22  &48.3       &53.5       &52.9           \\  
\hline
\end{tabular}
\end{table}

\noindent
\textbf{Tracking Speed.}
With the simplified and unified framework, our CEUTrack can not only realize SOTA performance on three frame-event tracking datasets but also boost high tracking efficiency. As demonstrated in Table~\ref{COESOT_auc}, our tracker achieves a speed of 75 FPS, ranking 3rd on the COESOT dataset and outperforming all of the multi-modal trackers, such as SiamR-CNN (5 FPS), LTMU~\cite{ltmu} (13 FPS), CMT-ATOM (14 FPS) and most of the other single-modal trackers. Note that although SiamRPN achieves the fastest tracking speed at 160 FPS, its accuracy significantly lags behind most trackers.

\begin{table}[t]
\centering
\small   
\caption{Ablation study of CEUTrack network in COESOT. } 
\label{component_ablation}
\begin{tabular}{l|c|ccccc}
\hline 
\textbf{Modules} & \textbf{ViT layers} & \textbf{SR}  &\textbf{PR} \\     
\hline
\textbf{w/o Pos\_Enc} & ViT-12   &49.0  &63.1 \\     
\textbf{w/o Adapters}   & ViT-12 & 62.0      & 74.7    \\     
\textbf{Adapters-4 } & ViT-12  & 62.5      & 75.1    \\     
\textbf{Adapters-6 }  & ViT-12  & 62.5      & 75.2   \\     
\textbf{Adapters-12 }  & ViT-12  & \textbf{62.7}      & \textbf{76.0}   \\     
Adapters-8 & \textbf{ViT-8  }    & 59.8       & 73.8  \\     
Adapters-4 & \textbf{ViT-4  }    & 47.6       & 61.9  \\     
\hline
\end{tabular}
\end{table}

\subsection{Ablation Study} 
\noindent
\textbf{Analysis on Network Components.} To ablate the usage of different components in CEUTrack, we report the tracking results on COESOT as shown in Table~\ref{component_ablation}. Without our adapter blocks in CEUTrack, the performance drops by 0.7\% on SR, 1.3\% on PR. Without positional encoding for our unified network, SR heavily drops to 49.0\%. With the same ViT-12 layers, we compared different adapter blocks (4, 6, and 12), and found that 12 adapter blocks contributed the most to our CEUTrack. Further, we compare different VIT layers in our feature extraction and it's easy to find the number of ViT Transformer blocks has a positive correlation with the performance (ViT-4$<$ViT-8$<$ViT-12).

\begin{table}[t]
\caption{Performance comparison with and without event data.} \label{event_data}
\begin{tabular}{l|ccccc}
\hline
\#. Input Data    &SR   & PR  & NPR & BOC \\
\hline
\text{1. Color Frames Only } & 58.9   & 71.6 & 71.0 & 18.1  \\
\text{2. Event Frames Only }  & 38.9   & 45.3 & 44.0  & 11.0 \\
\text{3. Event Voxels Only }  & 17.4   & 20.8 & 24.1   & 6.5   \\
\text{4. Frames + Event Voxels}  & \textbf{62.7}   & \textbf{76.0} &  \textbf{74.9}  & 21.3  \\
\hline
\end{tabular}
\end{table}

\noindent
\textbf{Analysis of Modality Complementary.} 
To validate the effectiveness of color frame and event stream usage for visual tracking, we report the tracking results on COESOT with unimodal and bimodal data. As shown in Table~\ref{event_data}, when only color frames are fed-in, we can get $18.1, 58.9, 71.6, 71.0$ on BOC, SR, PR, and NPR respectively.  While the overall tracking results can be improved significantly when the event streams are combined, i.e., $21.3, 62.7, 76.0, 74.9$. 
If we utilize event streams alone for tracking, the overall results are not satisfactory.  We can see that the event-based trackers using frames or voxels are all inferior to the tracker \#4. This could be because event streams are sparse in spatial view and difficult to distinguish from the background.
These results emphasize the importance of color frames and highlight the effectiveness of event streams, further validating the complementary nature of combining color and event data in visual tracking.


\begin{table}[t]
\centering
\small     
\caption{Performance comparison of event data formulation.} 
\label{input_ablation}
\setlength{\tabcolsep}{0.9mm}{
\begin{tabular}{l|ccccc}
\hline
~~~~Data Formulation    &SR  & PR   & NPR \\
\hline
\text{{1.} Input Fusion (Color + Event Frames) } & 59.0   & 72.0 & 71.7 \\
\text{{2.} Color Frames \& Event Frames }  & 60.4   & 73.7 & 72.5  \\
\text{{3.} Color Frames \& Event TimeSurface}& 60.9   & 74.2 & 72.8 \\
\text{{4.} Color Frames \& Event RecIMG}  & 60.5   & 73.8 & 72.4 \\
\text{{5.} Color Frames \& Event Voxel }  & \textbf{62.7}   & \textbf{76.0} &  \textbf{74.9}  \\
\hline
\end{tabular}
}  
\end{table}

\begin{figure*}[!htp]
\center
\includegraphics[width=1.0\textwidth]{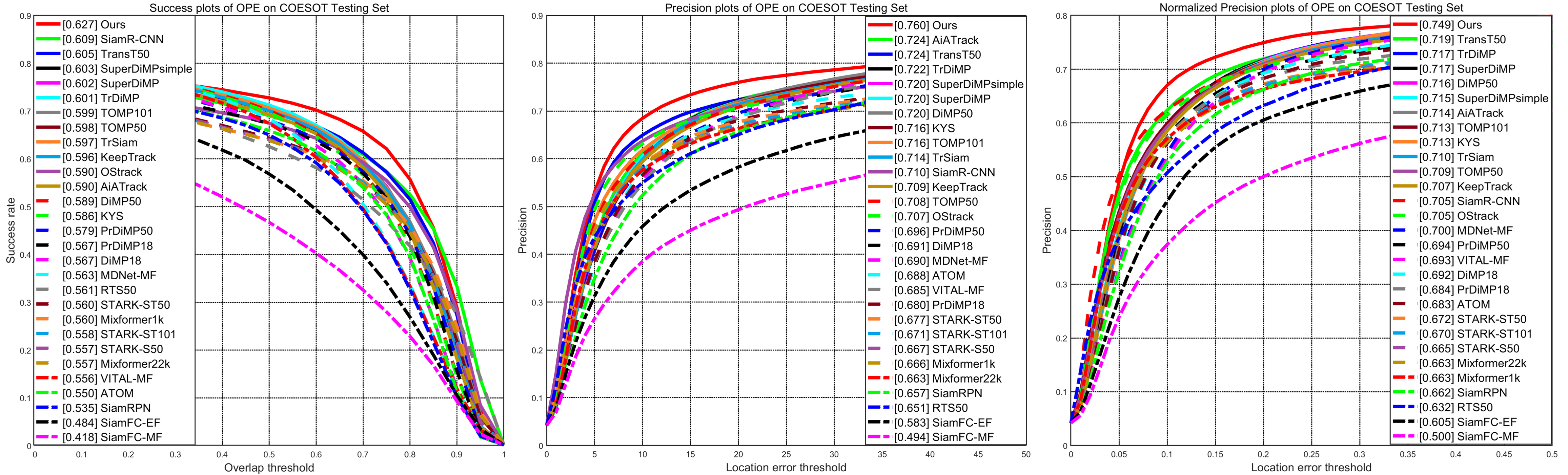}
\caption{Overall performance comparison on the proposed COESOT dataset. Best viewed by zooming in.}  
\label{COESOT}
\end{figure*}

\noindent 
\textbf{Analysis on Different Event Representations.} 
In this paper, we transform the event streams into voxel representation and achieve high-performance tracking ($62.7, 76.0, 74.9$ on SR, PR, NPR metrics), as the selected voxels consider the spatial sparsity and preserve the temporal information. Event streams, which capture temporal changes in the scene, are commonly represented in frame formats such as \textit{event frames}, \textit{timesurface}, and \textit{reconstructed grey images}. In this study, we train and test the tracking using these three representations and report the results in Table~\ref{input_ablation}.  When the color and event frames are combined at the image level, 
we obtain 59.0, 72.0, and 71.7 on the SR, PR, and NPR metrics, respectively.  However, by fusing the color and event frames at the feature level, we can improve the tracking results to 60.4, 73.7, and 72.5.  Further, we generate the timesurface representation for the event stream using the time-oriented approach~\cite{lagorce2016hots}, which achieves a performance of 60.9, 74.2, and 72.8 respectively.  We also reconstruct the grey image from event streams using the recurrent reconstruction model~\cite{rebecq2019events}, referred to as Event RecIMG in Table~\ref{input_ablation}. It achieves a similar performance of 60.5, 73.8, and 72.4 on the three metrics compared to other image-like event representations. From all the experimental results and comparisons, we observe that the combination of color frames and event voxels achieves the best tracking results on the COESOT dataset.

\noindent 
\textbf{Analysis on Voxel Sampling.}
Voxelization is a crucial component of our tracking framework, and in this subsection, we explore the influence of different sets of parameters for voxel sampling. Table~\ref{voxelnumber} shows the results of our tests, where we fix the spatial sampling at $m=\frac{W}{v_W}=34$ and $n=\frac{H}{v_H}=26$, and search for the appropriate values of $\tau=\frac{T}{v_T}$ and the number of voxels selected for the search and template. We find that the best results and an equilibrium speed can be achieved when $\tau=20$ and the top $4096/1024$ voxels are selected for the search/template.

\begin{table}
\centering
\caption{Performance comparison of extracted voxel number and top selection.} 
\label{voxelnumber}
\setlength{\tabcolsep}{0.9mm}{
\begin{tabular}{c|c|cccc}
\hline
 Voxel Numbers & Top Selection   & \multicolumn{3}{c}{Metrics}\\ 
 $Q$ $(m\times n\times \textbf{$\tau$})$        & (Search / Template)  & \text{SR}  &\text{PR}  &\text{BOC}  &\text{FPS}\\
\hline
\multirow{3}{*}{a. $34\times26\times \textbf{50}$} & \text{ 16384 / 4096} & 60.6 & 74.3 &18.9 & 65 \\
 &\text{ 4096 / 1024}  & 60.7 & 74.8 &19.3  & 75 \\
  &\text{ 1024 / 256} & 59.1 & 72.8 &18.5 & 78 \\
\hline
\multirow{3}{*}{b. $34\times26\times \textbf{20}$}  & \text{ 16384 / 4096} & 60.2 & 73.9  &19.4 & 65 \\
 & \textbf{ 4096 / 1024}  & \textbf{62.7} & \textbf{76.0}  &21.3  &\textbf{75} \\
 &\text{ 1024 / 256} & 59.7 & 73.2  & 19.0   & 78 \\
\hline
\end{tabular} 
} 
\end{table}

\begin{figure}
    \centering
    \includegraphics[width=\columnwidth]{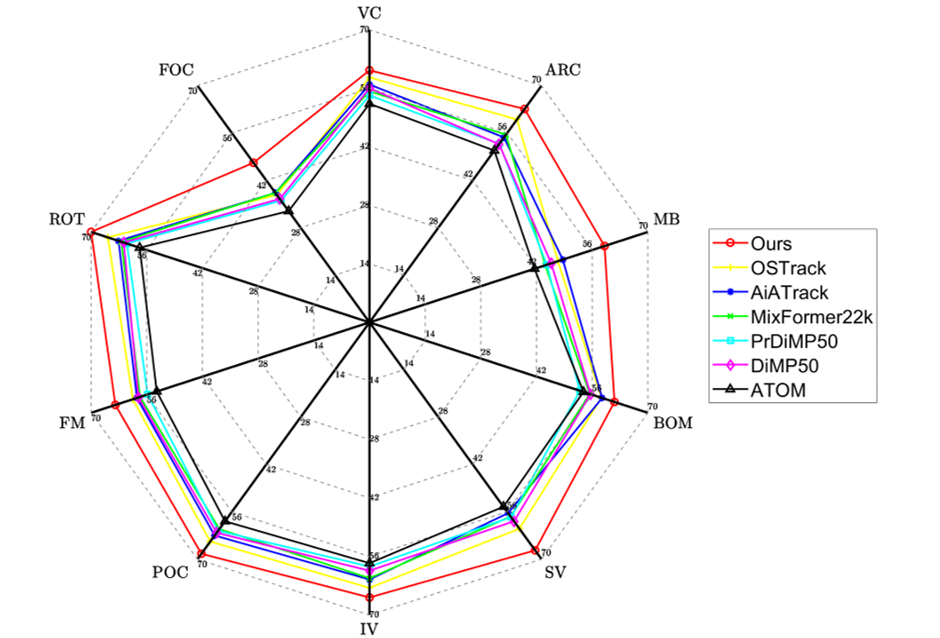}
    \caption{Success scores comparison of different attributes on COESOT.} 
    \label{radar_fig} 
\end{figure}


\begin{figure*}[!htp]
    \centering
\includegraphics[width=\textwidth]{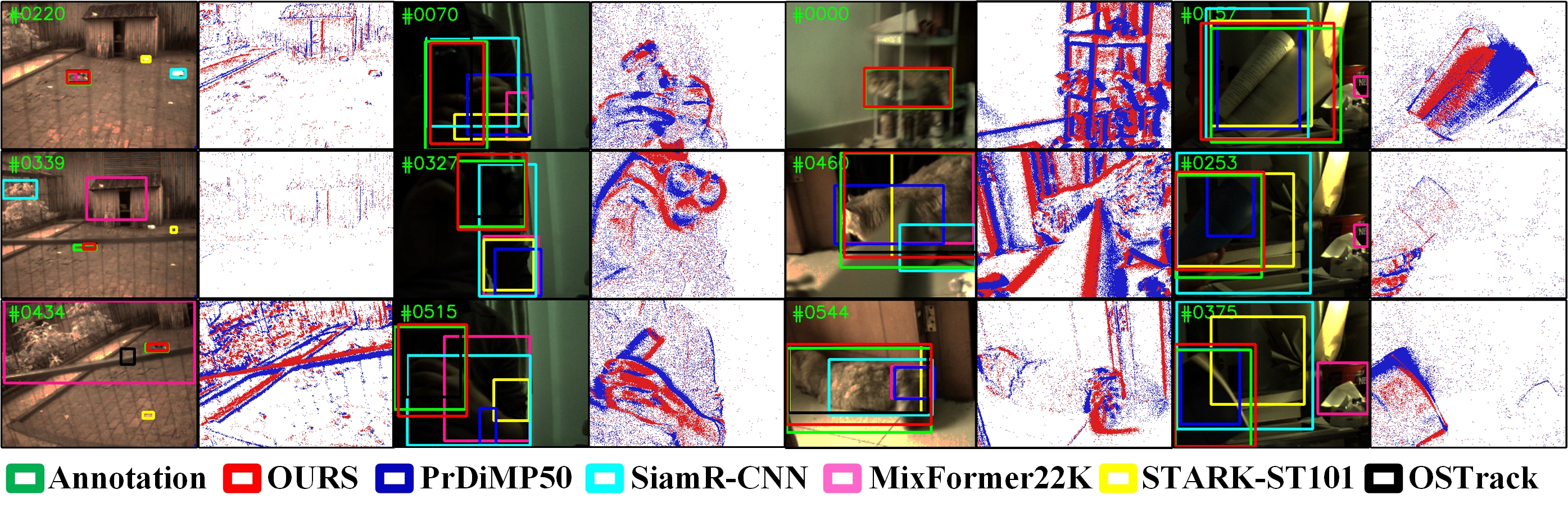}
    \caption{Visualization of tracking results on COESOT dataset. Event images are used for visual comparison only.} 
    \label{visTrackResults}
\end{figure*}

\subsection{Visualization and Analysis}
\noindent \textbf{Comparison Visualization.}
With the detailed attributes annotation of COESOT, we build the Matlab evaluation toolkit for plotting the diagram of curves about the overall performance and details attributes performance comparison. 
As shown in Fig.~\ref{COESOT}, the overall performance comparison of COESOT is reported about all of the baseline trackers and our CEUTrack obtains a considerable gain with all of them concerning SR/PR/NPR in 28 baseline trackers. 
We also evaluate the performance of each attribute and our CEUTrack achieves the best accuracy performance on almost all of these challenges, like motion blur, rotation,  fast motion, full occlusion, scale variation, etc. The state-of-the-art performance on Rotation, Fast Motion, and Motion Blur shows the contribution of event voxel modality. To aid in analyzing challenging factors, as shown in Fig.~\ref{radar_fig}, we report the radar results under 10 regular attributes, which can help developers compare the success rate easily and quickly. Our proposed CEUTrack outperforms SOTA trackers in addressing these challenging factors.

\begin{figure} 
    \centering
\includegraphics[width=1.0\columnwidth]{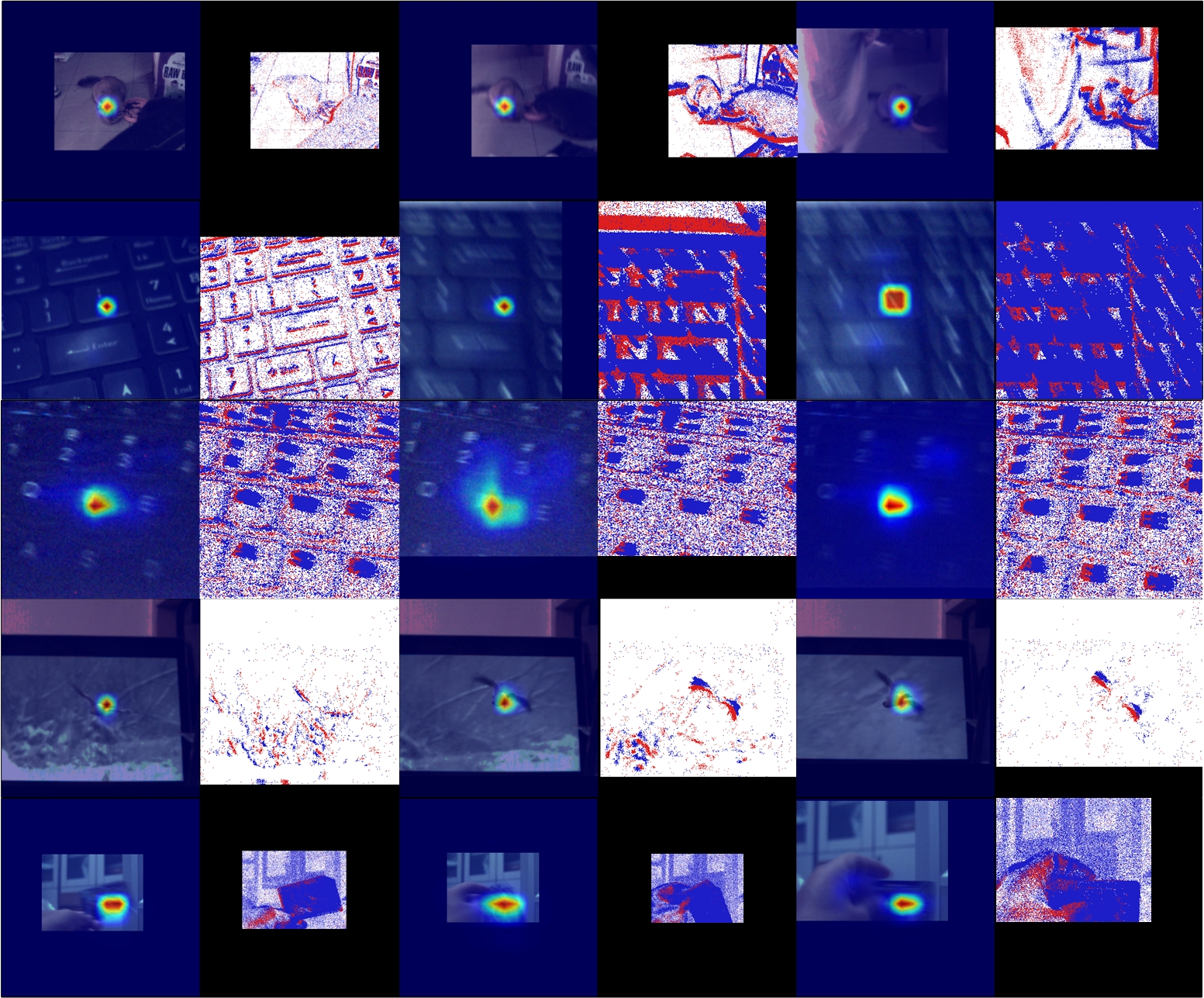}
    \caption{Visualization results of the response weight of our tracker on COESOT dataset.  Best viewed in color. }
    \label{feature_appendix}
\end{figure}

\noindent
\textbf{CEUTrack Visualization.} 
In addition to the detailed quantitative analysis provided in the previous sub-sections, we present visualizations of the feature maps and tracking results to facilitate a better understanding and qualitative analysis of our tracker. As illustrated in Fig.~\ref{visTrackResults}, our tracker (depicted by the red bounding box) exhibits robust and accurate tracking performance even under challenging conditions such as scale changes, low illumination, motion blur, and deformation, outperforming the current state-of-the-art trackers.
%


\noindent
 \textbf{Response Weight Visualization.}
We visualize many sequences of COESOT for the qualitative analysis of our unified backbone, as shown in Fig.~\ref{feature_appendix}. Due to the color and event frames being cropped 4 times region of the last prediction bounding box size, some outputs are padded with empty areas, represented by the black areas in visualization. As the second and third rows show, when fast motion and heavy motion blur challenges occurred, CEUTrack still obtained significant responses in the target object center. Further, when the videos suffer from rotation and view of changes (5th row), CEUTrack still responds well. The provided event frames are only for intuitive reference.

\begin{figure*}[!htp]
    \centering
    \includegraphics[width=\textwidth]{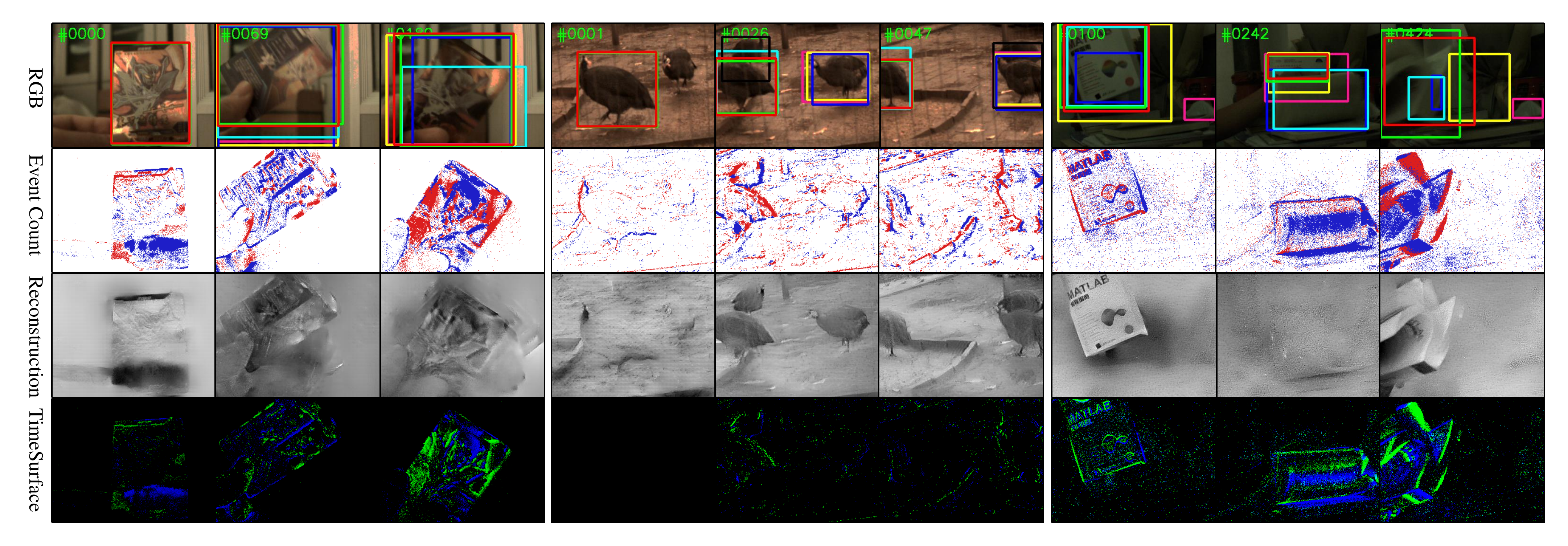}
    \caption{The color frames and three different event frames representation formats of COESOT.  Best viewed in color. }
    \label{event_representation}
\end{figure*}

\begin{figure}[!htp]
    \centering
\includegraphics[width=\columnwidth]{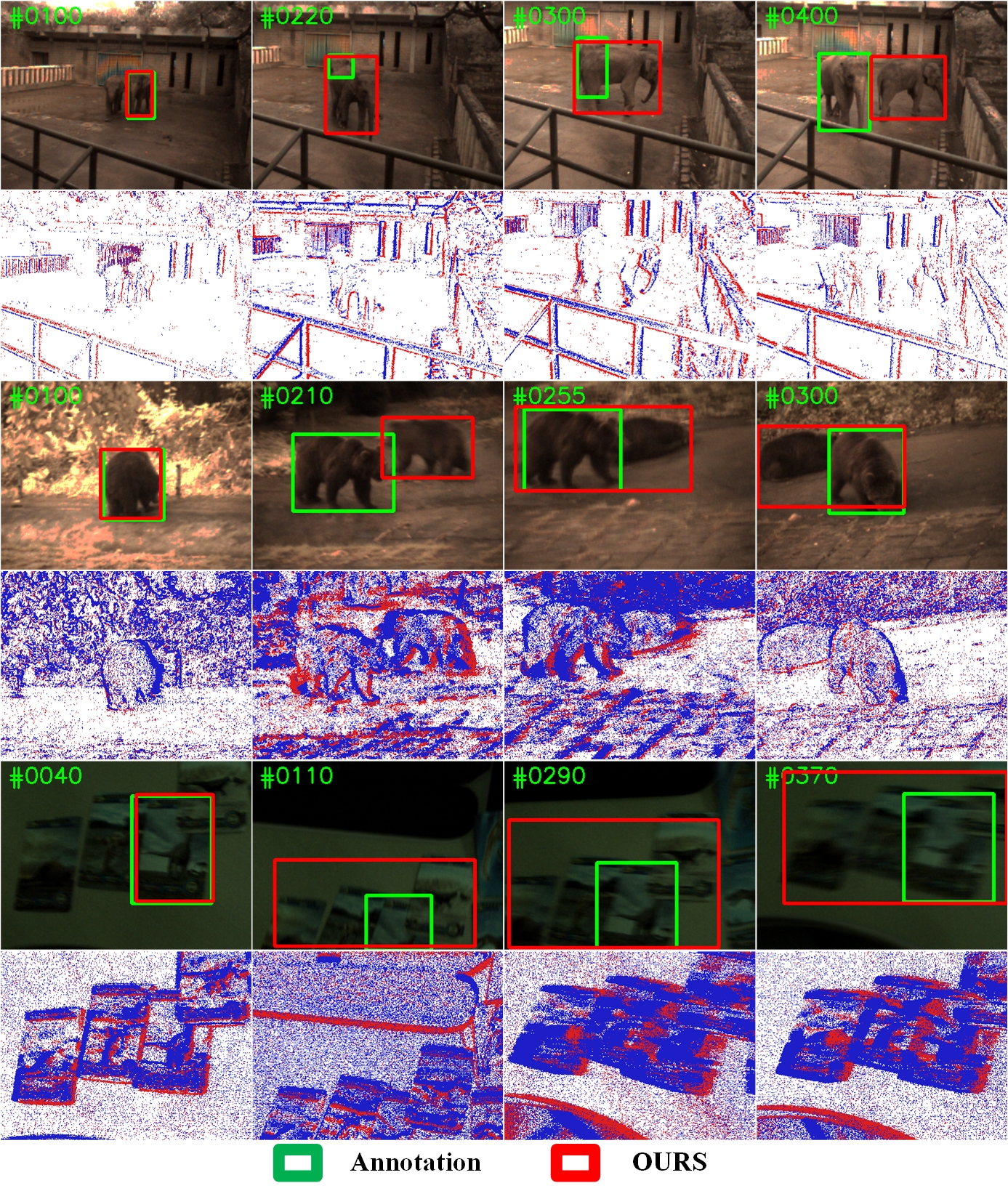}
    \caption{Visualization of failure cases of our CEUTrack. The event frames are provided by COESOT. Best viewed in color.}
    \label{fairlure}
\end{figure}

\noindent
\textbf{Event Representation Visualization.}
As the ablation study of event representation, the event can be represented by various methods, which can also be employed as reference input for future research. 
As shown in Fig.~\ref{event_representation}, we visualize three different event-building methods.
The first line is the color frames, and the second one is the most popular event count frames which add the event through the corresponding timestamps~\cite{wang2023visevent}. The third line is the event reconstruction frames, which build event frames from a stream of events based on a recurrent network~\cite{rebecq2019events}. The last one is the time-surfaces event presentation formulation~\cite{lagorce2016hots}, which describes event-based spatiotemporal features.
In our proposed COESOT dataset, for all video sequences, we have provided color frames, the above three event formats, and our event voxel representation format for researchers to choose from.
Corresponding conversion scripts are also provided for an easy start. Further, we provide the original source file with the \textit{.aedat4} extension for all video sequences to facilitate future research, e.g., transforming them into new event representations.

\subsection{Failure Cases and Limitation Analysis}
\noindent We believe the analysis above can prove the contribution of our tracker and dataset. At the same time, our tracker has some failure cases as shown in Fig.~\ref{fairlure}.
In particular, it shows four frames of three different COESOT sequences containing ground truth and our CEUTrack prediction bounding boxes.  To sum up, CEUTrack fails when similar distractors are close to the target due to the lack of an anti-distractor mechanism. When the distractor partially or heavily occluded the target, the tracker mistakenly tracked the similar distractor (first sequence in Fig.~\ref{fairlure}). Another similar scenario is where the target and distractor approach each other (second sequence), and CEUTrack covers both the target object and the distractor. Once the two objects diverge again, the tracker will track the object or the distractor with a similar probability. The last failure scenario is the multiple similar distractors with fast camera irregular motion, which can't predict the target location from dense objects, and therefore covering the whole object position as the tracking result. Therefore, augmenting the anti-distractors mechanism is a very helpful method to address the lack of semantic difference between target and distractors in event modality.    

In real tracking scenarios, when the target remains stationary (No Motion) or moves very slowly under dark/overexposed, our method may be violated due to too sparse dynamic event voxels (or no existing events) and limited color view. 
Further, fast camera movement will generate too dense events and the voxel top-selection strategy may extract not enough search voxels, indicating that there is still unexplored room for color-event based object tracking.

\section{Conclusion} 
In this paper, we present a unified network CEUTrack, which capitalizes voxelization to adeptly capture spatial-temporal details from event streams and integrates the multi-model tracking within an adaptive single-stream network, obviating the intricate fusion design of modern multi-model tracking methods. 
Extensive experiments on datasets substantiate the effectiveness and efficiency of our tracker. 
To address the scarcity of current benchmarks and stimulate further research in color-event tracking, we propose a large-scale and high-quality dataset COESOT. 
Additionally, we provide a toolkit tailored for tracking evaluation and comparison, which encompasses 28 baseline tracking results on COESOT. We newly introduce a metric BOC for validating the real effectiveness and conspicuousness ability compared to existing trackers. In future work, we contemplate pre-training our backbone using paired frame and event streams in a self-supervised manner to achieve higher tracking performance.


\section*{Data Availability Statement}
\noindent
The authors confirm the data supporting the findings of this work are available within the article or its supplementary materials.

\bibliography{bibliography}

\end{document}